\documentclass{sig-alternate-2013}

\permission{Copyright is held by the International World Wide Web Conference Committee (IW3C2). IW3C2 reserves the right to provide a hyperlink to the author's site if the Material is used in electronic media.}
\conferenceinfo{WWW'15 Companion,}{May 18--22, 2015, Florence, Italy.} 
\copyrightetc{ACM \the\acmcopyr}
\crdata{978-1-4503-3473-0/15/05. \\
http://dx.doi.org/10.1145/2740908.2742570}

\usepackage{algorithm,algpseudocode}
\usepackage{subfig}

\begin{document}

\title{Detecting Singleton Review Spammers Using Semantic Similarity}

\numberofauthors{2} 
\author{
\alignauthor
Vlad Sandulescu\titlenote{This paper extends some of the results from the author's MSc thesis (Unpublished) \cite{sandulescu:opinion}. Research done while working at Trustpilot.}\\
	\affaddr{Adform}\\
       \affaddr{Copenhagen, Denmark}\\
       \email{vlad.sandulescu@gmail.com}
\alignauthor
Martin Ester\\
       \affaddr{School of Computing Science}\\
       \affaddr{Simon Fraser University}\\
       \affaddr{Burnaby, BC, Canada}\\
       \email{ester@cs.sfu.cs}
}

\maketitle
\begin{abstract}
Online reviews have increasingly become a very important resource for consumers when making purchases. Though it is becoming more and more difficult for people to make well-informed buying decisions without being deceived by fake reviews. Prior works on the opinion spam problem mostly considered classifying fake reviews using behavioral user patterns. They focused on prolific users who write more than a couple of reviews, discarding one-time reviewers. The number of singleton reviewers however is expected to be high for many review websites. While behavioral patterns are effective when dealing with elite users, for one-time reviewers, the review text needs to be exploited. In this paper we tackle the problem of detecting fake reviews written by the same person using multiple names, posting each review under a different name. We propose two methods to detect similar reviews and show the results generally outperform the vectorial similarity measures used in prior works. The first method extends the semantic similarity between words to the reviews level. The second method is based on topic modeling and exploits the similarity of the reviews topic distributions using two models: bag-of-words and bag-of-opinion-phrases. The experiments were conducted on reviews from three different datasets: Yelp (57K reviews), Trustpilot (9K reviews) and Ott dataset (800 reviews). 
\end{abstract}


\section{Introduction}
The number of consumers that first read reviews about a product they wish to buy is constantly on the rise.
As consumers increasingly rely on these ratings, the incentive for companies to try to produce fake reviews to boost sales is also increasing. Spam reviews have at least two major damaging effects for the consumers. First, they lead the consumer to make bad decisions when buying a product. After reading a bunch of reviews, it might look like a good choice to buy the product, since many praised it. After buying it, it turns out the product quality is way below expectations and the buyer is disappointed. Second, the consumer's trust in online reviews drops.

There are two directions where the research on opinion spam has focused on so far: behavioral features and text analysis. Behavioral features represent things like the review rating, review date, IP from where the review was posted and so on. Textual analysis refers to methods used to extract clues from the review content, anywhere from parts-of-speech patterns to word frequency. The behavioral models have shown good standalone results, while linguistic models, based on cosine similarity or n-grams, were less precise, although they did bring small improvements to the overall model accuracy when added on top of the behavioral features, see \cite{feng:distributional, jindal:opinion, lim:detecting, mukherjee:spottingopinion, mukherjee:spottingfake, mukherjee:fake, mukherjee:what, ott:finding}. The authors of \cite{zengin:user} concluded that human judgment used to detect semantic similarity of web document does not correlate well with cosine similarity.

Most studies consider only users who write at least a couple of reviews and one-time reviewers are eliminated from the test datasets. However, a very large proportion of reviewers is assumed to only post a single review under a single user name. This assumption is based on studies which used real-life commercial reviews, such as \cite{xie:review}, who observed that over 90\% of the reviewers of resellerratings.com only wrote one review. It is also strongly confirmed by the author's experience while employed at Trustpilot. For one time reviewers, behavioral clues are scarce, so in this paper we argue the key to catch this type of spammers can only be found in the review text. We make an important assumption, also noted in \cite{ott:finding}: spammers have a limited imagination when it comes down to writing completely new details in every review. They are prone to rephrasing, switching some words with their synonyms while keeping the overall review sentiment the same. We argue that semantic similarity can capture more subtle textual clues that make several reviews written under different names to point to a single person.

This paper proposes two approaches to detect fake reviews based only on their text and makes the following contributions:

1. It proposes a method which uses the knowledge-based semantic similarity measure described in \cite{mihalcea:corpus}. This exploits the synonymy relations between words, contained in the synsets of WordNet and devises a weighted formula to compute the similarity between any two documents. We create extensions of the cosine similarity measure through extracting only specific parts-of-speech (POS) patterns from a document and by employing lemmatization. The vectorial-based measures are used as baselines and the results of the semantic measure is compared against them. To our knowledge, this is the first study to combine semantic similarity on top of the popular WordNet synonyms database for the purpose of fake review detection. 

2. It proposes a detection method which draws from models aimed at extracting product aspects from short texts, such as user opinions and forums. In recent years, topic modeling and in particular Latent Dirichlet Allocation (LDA) have been proven to work very well for this problem. The novelty of the proposed method is to use the similarity of the underlying topic distributions of reviews to classify them as truthful or spam. We used two models, a bag-of-words which included a restricted set of POSs and a bag-of-opinion-phrases model described in \cite{moghaddam:aspect}. The latter splits a review into aspect-sentiment pairs, and these pairs are then used in the LDA model, instead of the document words. To our knowledge, the latter model has never been used in conjunction with fake reviews detection.

3. It conducts a thorough experimentation to evaluate the proposed models using 3 reviews datasets. The results are compared to the baselines and it is argued they generally perform better. This represents a strong indicator of the generalization power of the models in real life scenarios. To our knowledge, no fake reviews detection method has been tested on such a wide variety of reviews. 

\section{Related Work}

The opinion spam problem was first formulated by Jindal and Liu in the context of product reviews \cite{jindal:opinion}. By analyzing Amazon data and using near-duplicate reviews as positive training data, they showed how widespread the problem of fake reviews was at that time.

The first study to tackle the opinion spam as a distributional anomaly was described in \cite{feng:distributional}. It claimed product reviews are characterized by natural distributions which are distorted by hired spammers when writing fake reviews. They conducted a range of experiments that found a connection between distributional anomalies and the time windows when spam reviews were written. 

Another detection method which relied on behavioral user features and discarded singleton reviewers is described in \cite{fei:exploiting}. The focus was on supervised classification of spammers who write reviews in short bursts, using a graph propagation method in the reviewers graph. This method, as well as the ones described in \cite{lim:detecting, mukherjee:spottingopinion, mukherjee:spottingfake} can only detect fake reviews written by elite users on a review platform but exploiting review posting bursts is an intuitive way to obtain smaller time windows where suspicious activity occurs, similar to \cite{feng:distributional}. 

The authors of \cite{ott:finding} employed crowdsourcing through the Amazon Mechanical Turk (AMT) to create a gold-standard fake reviews dataset. The study proved humans cannot distinguish fake reviews by simply reading the text, showing an at-chance probability. It used an n-gram model and coupled the most frequent POS tags in the review text with psycholinguistic features. The experiments also revealed words associated with imaginative writing. While the classifier scored good results on the AMT dataset, it can be argued that once the spammers learn about these, they could simply avoid using them. The ability of models evaluated only against the artificially obtained AMT dataset have been proven to not generalize to real life scenarios, as proven in \cite{mukherjee:fake}. This study attempted to recreate the model of \cite{ott:finding} and evaluate it on Yelp reviews, assuming that Yelp had perfected its fraud detection algorithms in its decade of existence. Surprisingly the results showed only a 68\% accuracy when they tested Ott's model on Yelp data instead of the 90\% accuracy reported on the AMT data.

\cite{mukherjee:spottingfake} were the first to try to solve the problem of opinion spam resulted from a group collaboration between multiple spammers. The study used frequent itemset mining to build candidate groups of users who posted more than a couple of reviews together for the same businesses. The authors ranked and classified spammers based on weighted spamicity models built from individual and group behavioral indicators, e.g. rating deviation between group members, number of products the group members worked together on, or review content overlap using cosine similarity. The evaluation dataset was built using review annotation by human judges. The algorithm considerably outperformed existing methods by achieving an area under the curve result (AUC) of 95\%. In \cite{mukherjee:spottingopinion}, the same authors built an unsupervised model which used the same behavioral features from \cite{mukherjee:spottingfake, mukherjee:fake} and exploited the distributional divergence between honest users and spammers in terms of their behavioral footprints. The novelty about the proposed method in this paper is a posterior density analysis of each of the features used. 
In \cite{mukherjee:what} the authors made an interesting observation in their study: the spammers caught by Yelp's filter seem to have "overdone faking" in their attempt to sound more genuine. In their spam reviews, they tried to use words that appear in genuine reviews almost equally frequent, thus avoiding to reuse the exact same words in their reviews. This supports the claim that cosine similarity is not enough to catch more subtle spammers in real life scenarios.

The only study which specifically targets singleton reviewers is \cite{xie:review}. The authors observed over 90\% of the reviewers of resellerratings.com only wrote one review, so they have focused their research on this type of reviewers. They also claim, similarly to \cite{feng:distributional}, that a flow of fake reviews coming from a hired spammer distorts the usual distribution of ratings for a product. They observed bursts of either very high or very low ratings, so they tried to detect time windows in which these abnormally correlated patterns, such as number of reviews, average ratings and the ratio of singleton reviews appear.

\section{Models and experimental setup}
In this section, we first give an overview of the vectorial, semantic similarity and LDA models and describe the actual similarity measures we used. Then we present the evaluation datasets and the review text preprocessing steps for the chosen models.

\subsection{Vectorial and semantic similarity measures}

Textual similarity is ubiquitous in most natural language processing problems. Computing the similarity of text documents boils down to computing the similarity of their words. Any two words can be considered similar if some relation can be defined between them: e.g. synonymy/antonymy, or they might be used in the same context (legislation and parliament), or they could both be nouns or verbs.

The vector space model is widely used in information retrieval to find the best matching set of relevant documents for an input query. Given two documents, each of them can be modeled as a n-dimensional vector where each of the dimensions is one of its words. Their similarity is computed by measuring the cosine angle between their two word-vectors.

For two vectors $T_1$ and $T_2$, their cosine similarity is formulated as
\begin{equation}
\cos ({ T_1},{ T_2})= {{T_1} {T_2} \over \|{ T_1}\| \|{ T_2}\|} = \frac{ \sum_{i=1}^{n}{{ T_1}_i{ T_2}_i} }{ \sqrt{\sum_{i=1}^{n}{({ T_1}_i)^2}} \sqrt{\sum_{i=1}^{n}{({ T_2}_i)^2}} }
\end{equation}

Although simple yet effective, the vector based model has some disadvantages. It assumes the query words are statistically independent, but the document would not make any sense if this were true. It captures no semantic relationships, so documents from the same context with words from different vocabularies are not seen as similar. Various improvements to this model, such as stopwords removal or POS tagging have been proposed over time.

According to \cite{budanitsky:lexical}, there are a few notable approaches to measure semantic relatedness and many of them are based on WordNet. It is one of the largest lexical databases for the English language, comprising of nouns, verbs, adjectives and adverbs grouped in synonymical rings, i.e. groups of words, which in the context of information retrieval applications are considered semantically equivalent. These groups are then connected through relations into a larger network which can be navigated and used in computational linguistics problems. 

The authors of \cite{mihalcea:corpus} proposed a method to extend the word-to-word similarity inside WordNet to the document level and proved their method outperformed the existing vectorial approaches. 
They combined the maximum value from six well known word-to-word similarity measures based on distances in WordNet, with the inverse document frequency metric \textit{idf} for a word \textit{w}. 
They formulated the semantic similarity between two texts $T_1$ and $T_2$ as

\begin{equation}\label{eq:mihalcea-definition}
\begin{split}
sim ({ T_1},{ T_2})= \frac{1}{2}(\frac{\sum\limits_{w\in\{T_1\}}{(maxSim(w,T_2)*idf(w))}}{\sum\limits_{w\in\{T_1\}}{idf(w)}} + \\ \frac{\sum\limits_{w\in\{T_2\}}{(maxSim(w,T_1)*idf(w))}}{\sum\limits_{w\in\{T_2\}}{idf(w)}})
\end{split}
\end{equation}
We have also experimented with improvements of the simple cosine similarity measure:\\
$\bullet$ cosine similarity with all POS tags, excluding stopwords (baseline)\\
$\bullet$ cosine similarity with non-lemmatized POSs - nouns, verbs and adjectives, excluding stopwords\\
$\bullet$ cosine similarity with lemmatized POSs - nouns, verbs and adjectives, excluding stopwords\\
$\bullet$ mihalcea semantic similarity - nouns, verbs and adjectives, excluding stopwords\\
We have considered the results of the cosine similarity measure applied to a documents pair after removing the stopwords as baseline. 

The pseudocode of the algorithm used to compute the pairwise similarity between reviews for a particular business is detailed in Algorithm \ref{alg:singleton}.

\begin{algorithm}
\caption{Pseudocode for the detection method using semantic similarity}
\label{alg:singleton}
\begin{algorithmic}[1]
\For{each review $R$ in dataset}
	\State Remove stopwords
  	\State Extract POSs (nouns, verbs and adjectives)
  	\EndFor
  	\For{each business $B$}
  		\For{each reviews pair ($R_i$, $R_j$) $\in$ $B$}
		\State $\underset{R_i, R_j \in B}{sim}(R_i, R_j)$ = $\underset{R_i, R_j \in B}{similarity\_measure}(R_i, R_j)$
	\newline\Comment{${similarity\_measure}$ is each measure out of \{cosine, cosine pos non lemmatized, cosine pos lemmatized and mihalcea\}}
		\EndFor
		\For{$spam\ threshold\ T = 0.5$, $T <= 1$, $T{+=}{0.05}$}
			\If{${sim(R_i, R_j)}$ > ${T}$}
				\State Mark $R_i$ and $R_j$ as spam
			\Else 
				\State Mark $R_i$ and $R_j$ as truthful
			\EndIf
		\EndFor
	\EndFor
\end{algorithmic}
\end{algorithm}

\subsection{LDA model}
Topic models are statistical models where each document is seen as a mixture of latent topics, each of the topics contributing with certain proportions to the document. They are explained more thoroughly in \cite{blei:latent}. These models have received increasing attention since they do not require manually labeled data to work, although they perform best when trained on large datasets. 

Reviews are in fact short documents and can be abstracted as a mixture of latent topics. The topics can be equivalent to the review aspects, so extracting aspects can become a topic modeling problem. It should then be possible to detect opinion spam, by comparing the similarity of the underlying topic distributions of review aspects with a fixed spam threshold. As \cite{moghaddam:aspect} notes, the topics may refer to both aspects - \textit{laptop}, \textit{screen} and sentiments - \textit{expensive}, \textit{broken}. Several LDA-based models have been proposed by \cite{moghaddam:onthedesign} which also evaluated which technique, frequent nouns, POS patterns or opinion phrases (<aspect,sentiment> pairs) performs best.

The Kullback-Leibner (KL) measure can be used to compute the difference between two probability distributions \textit{P} and \textit{Q}. But the measure is undefined if \textit{Q(i)} is zero and also not symmetric, meaning the divergence from \textit{P} to \textit{Q} is not the same as that from \textit{Q} to \textit{P}.
The Jensen-Shannon (JS) measure addresses both these drawbacks. It is also bounded by 1, which is more useful when comparing a similarity value for a review pair with a fixed threshold in order to classify the reviews as fake. Equation \ref{eq:JS} formulates the JS measure.
\begin{equation}\label{eq:JS}
\begin{split}
JS(P \parallel Q)= \frac{1}{2}KL(P \parallel M)+\frac{1}{2}KL(Q \parallel M), \\ where \hspace{0.2cm} M=\frac{1}{2}(P+Q)
\end{split}
\end{equation}
The JS measure can be rewritten in the form of equation \ref{eq:IR}, in order to decrease computational time for large vocabularies, as mentioned by \cite{dagan:similarity}. IR is short for information radius, while {$\beta$} is a statistical control parameter.
\begin{equation}\label{eq:IR}
IR(P,Q) = 10^{-\beta{JS(P \parallel Q)}}
\end{equation}
We computed the pairwise IR similarity value, for all reviews of a business, which remained after the preprocessing step and for several number of topics $\in$ \{10, 30, 50, 70, 100\}. This was then compared to a fixed spam threshold in order to predict whether the reviews were truthful or spam,  similar to the semantic approach described in Algorithm \ref{alg:singleton}.

\subsection{Datasets and text preprocessing}
We crawled Yelp and built a dataset of 57K reviews from 660 New York restaurants and considered Yelp's recommended reviews (unfiltered) as truthful and the not recommended (filtered) ones as spam. Several well known studies have considered Yelp's filtered reviews as fake and unfiltered ones as truthful \cite{mukherjee:fake, mukherjee:what}. We balanced out the Yelp dataset, such that each business would have an equal number of recommended and not recommended reviews. The Trustpilot dataset of 9K labeled English reviews was kindly shared with us by the company. It contains 4 and 5 star reviews from 130 businesses, from one-time reviewers only. It is already balanced between truthful and fake reviews. The company has been filtering away fake reviews for several years now, so we have assumed their detection mechanisms provide fairly good results. The Ott dataset contains 800 reviews, balanced between truthful and fake and is publicly available \cite{ott:finding}. The dataset was created through AMT crowdsourcing, by soliciting participants to pretend working in the marketing department of hotels and write fake reviews for their employers. The authors allowed only one submission per turker account and rejected short, illegible or plagiarized reviews. 

Several steps were needed in order to get the review text in a processable shape and to compute the vectorial and semantic similarity between reviews. We removed stopwords from all the reviews and used a POS tagger \cite{toutanova:feature} to tokenize the reviews and extract only some POSs for further processing.

For the LDA models, besides the standard preprocessing, we checked the word frequency distribution of the remaining words. The goal was to remove more uninformative, highly seller-specific words, words with very low frequency (did not appear at least twice), as well as highly frequent words (appeared more than 100 times). The frequencies were empirically adjusted. These words were inducing noise into the topic distribution and could also be added to the stopwords list from the initial preprocessing step. Furthermore, reviews which did not have at least 10 words after the frequency-based filtering step were also removed from the corpus, since the quality of the LDA topics is known to be low for short and sparse texts and the similarity score of a sparse review pair would not be very relevant.
 
\section{Results}
\subsection{Semantic similarity - Yelp}
\begin{figure}
\centering
\subfloat[Subfigure 1 list of figures text][Precision]{
\includegraphics[width=0.23\textwidth]{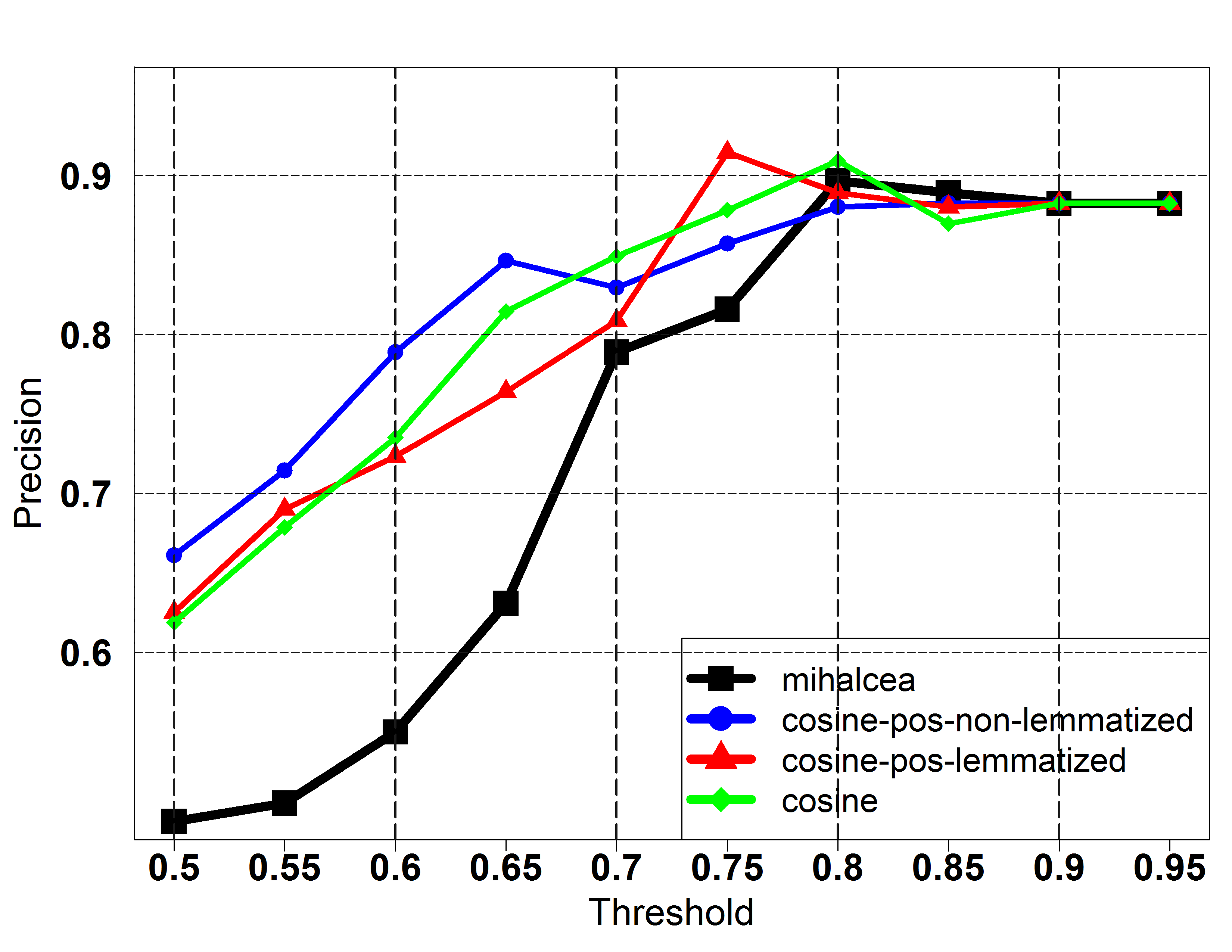}
\label{fig:yelp-semantic-precision}}
\subfloat[Subfigure 2 list of figures text][F1 score]{
\includegraphics[width=0.23\textwidth]{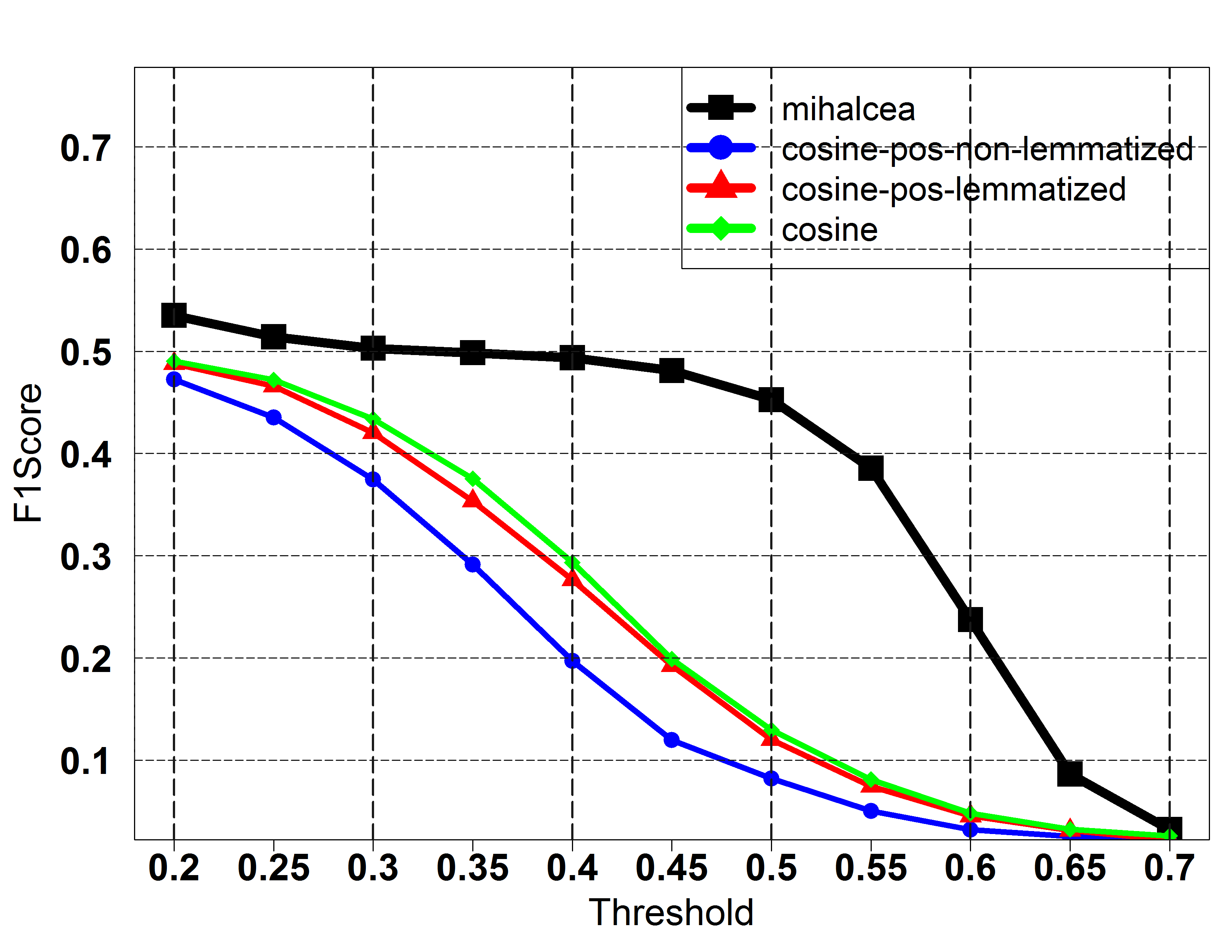}
\label{fig:yelp-semantic-f1score}}
\caption{Yelp - classifier performance with vectorial and semantic similarity measures}    
\label{fig:semantic_yelp}
\end{figure}

Figure \ref{fig:semantic_yelp} shows the classifier performance for Yelp reviews using the cosine measure (green), its variations (including and excluding lemmatization) and the semantic measure (black). Out of the vectorial measures, the cosine similarity with restricted POS tags and lemmatization achieves the highest precision for a threshold of 0.75. The intuition that the scores should become more precise as the threshold is raised is proven by the results. Precision is 90\% when the similarity threshold equals 0.8, as shown in Figure \ref{fig:yelp-semantic-precision}. The semantic measure is generally very close and achieves higher precision than the vectorial measures above a threshold of 0.8. Above such thresholds the recall is low, making the overall F1 score plotted in Figure \ref{fig:yelp-semantic-f1score} consequently low. However, overall the semantic measure outperforms the vectorial baselines. 

\subsection{Semantic similarity - Trustpilot}
\begin{figure}
\centering
\subfloat[Subfigure 1 list of figures text][Precision]{
\includegraphics[width=0.23\textwidth]{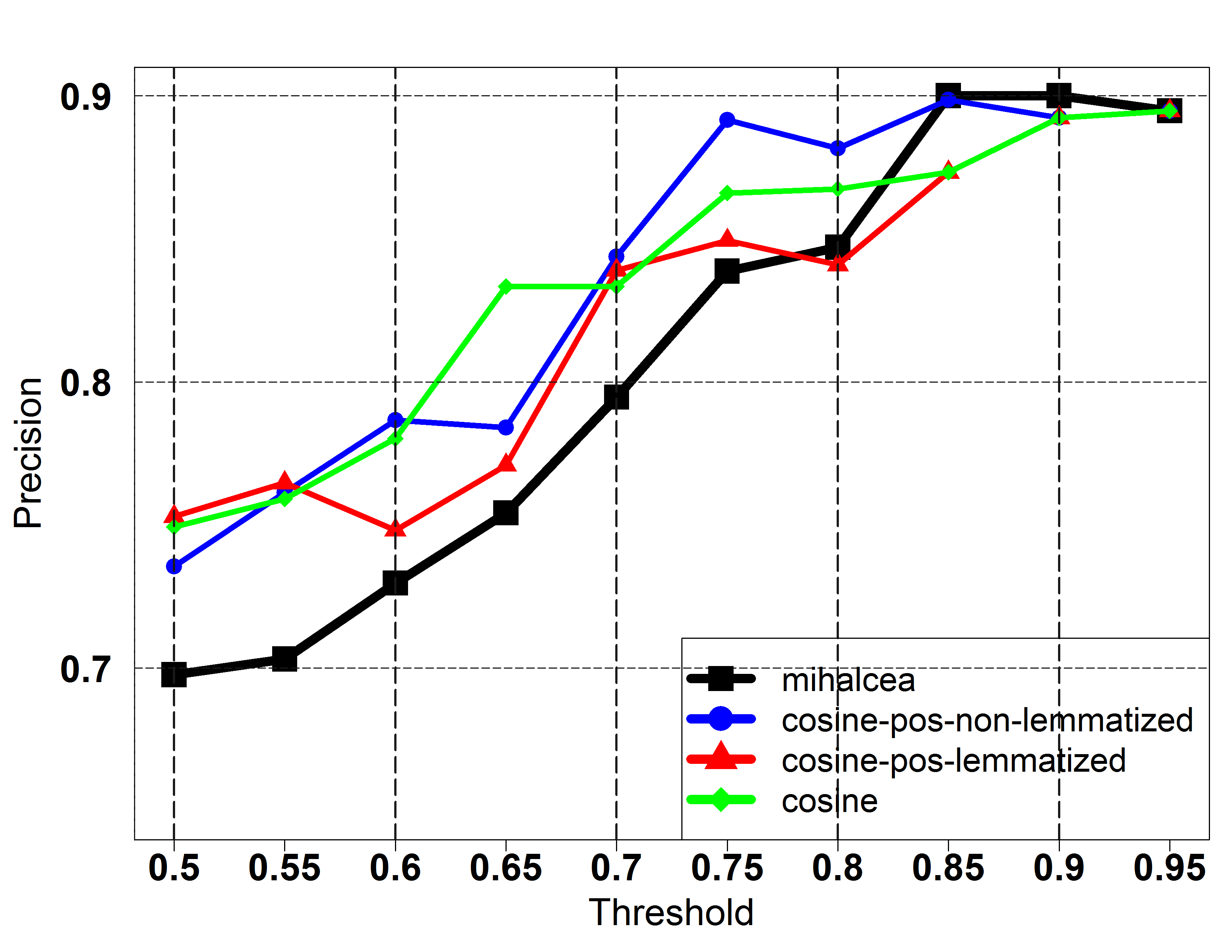}
\label{fig:trustpilot-semantic-precision}}
\subfloat[Subfigure 2 list of figures text][F1 score]{
\includegraphics[width=0.23\textwidth]{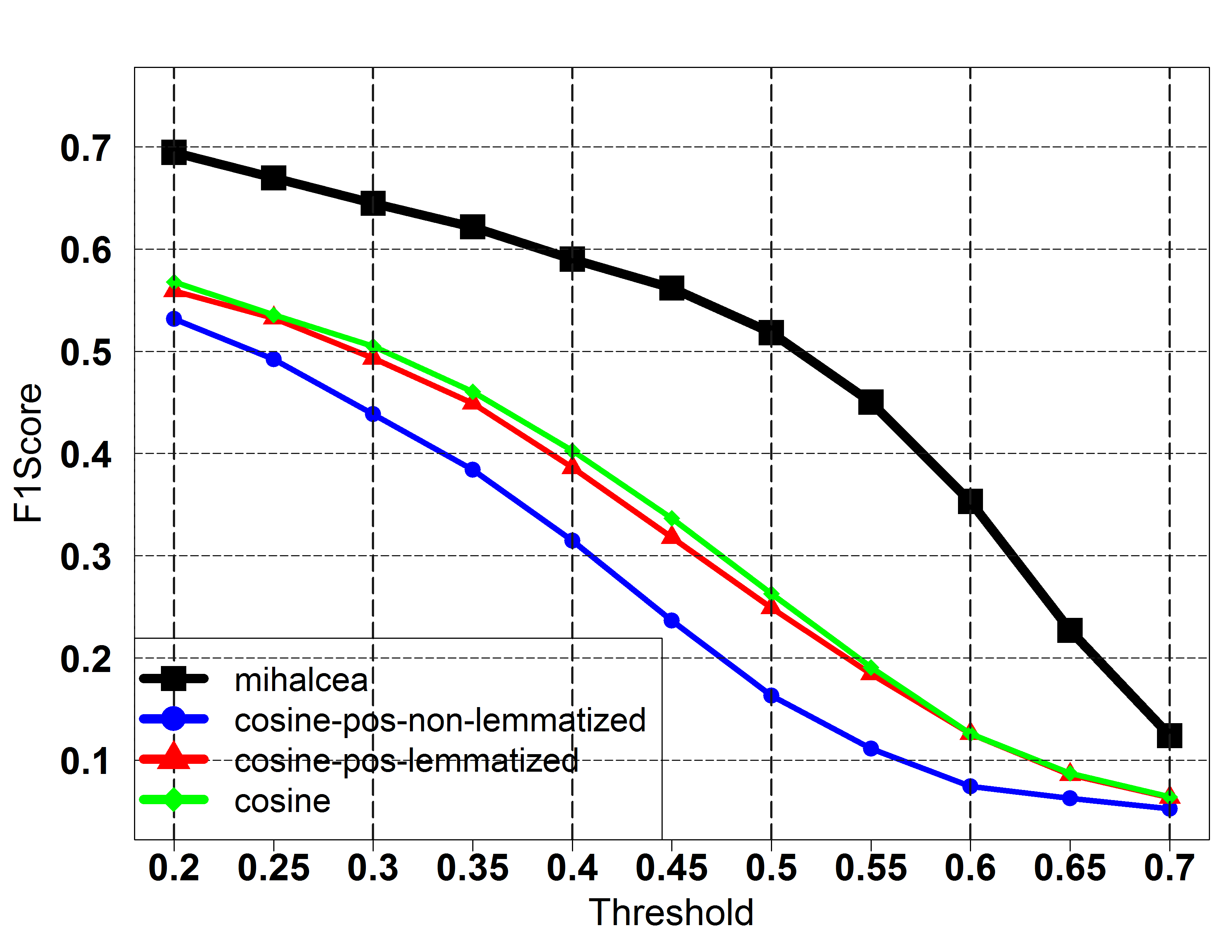}
\label{fig:trustpilot-semantic-f1score}}
\caption{Trustpilot - classifier performance with vectorial and semantic similarity measures}    
\label{fig:semantic_trustpilot}
\end{figure}

The evaluation of the classifier performance on the Trustpilot dataset is shown in Figure \ref{fig:semantic_trustpilot}. The semantic similarity method again showed better overall results compared to the cosine baselines. Its precision follows a smoother climb towards higher thresholds and follows the cosine baselines closer, compared to the Yelp reviews. The recall for the semantic method is also noticeably higher than the other methods. Precision of over 80\% is reached for similarity thresholds of above 0.7 and goes over 90\% above 0.85 threshold. Although the cosine with lemmatization performed better than without lemmatization, it sensibly achieved a lower precision.

One possible explanation to why the recall and F1 score are higher for Trustpilot than for Yelp might be that the opinion spammers targeting Trustpilot are not that professional. They do not make the effort to write more elaborate reviews, mimicking the honest reviewers writing styles. They seem much more prone to reuse the same exact words or synonyms when writing new reviews. Another explanation could lie in the recommended/not recommended reviews product feature of Yelp and the fact that some of the reviews which end up not being recommended might not be fake per se, but only less informative. Thus spammers have to first pass the "content recommendation" filter and write a meaty review with enough text to get it published. This needs considerably more effort and so it is more likely they will take the extra step to blend in better with the honest users, deceiving even the semantic similarity approach. It appears the Yelp spammers are doing a good job blending in with the honest reviewers, as it was also signaled in \cite{mukherjee:what}.

\subsection{Distribution of reviews in Ott dataset}
The purpose of this research paper is to detect fake reviews written by the same person under multiple names, therefore we did not feel it would make sense to run the classifier based on semantic similarity on the Ott dataset due to the way it was built. However, since the paper \cite{ott:finding} has received considerable citations, we were curious whether semantic similarity and cosine similarity derivatives would capture differences in the two review distributions across the entire dataset.

\begin{figure}
\centering
\subfloat[Subfigure 1 list of figures text][Mihalcea]{
\includegraphics[width=0.23\textwidth, trim = 9mm 15mm 8mm 15mm, clip]{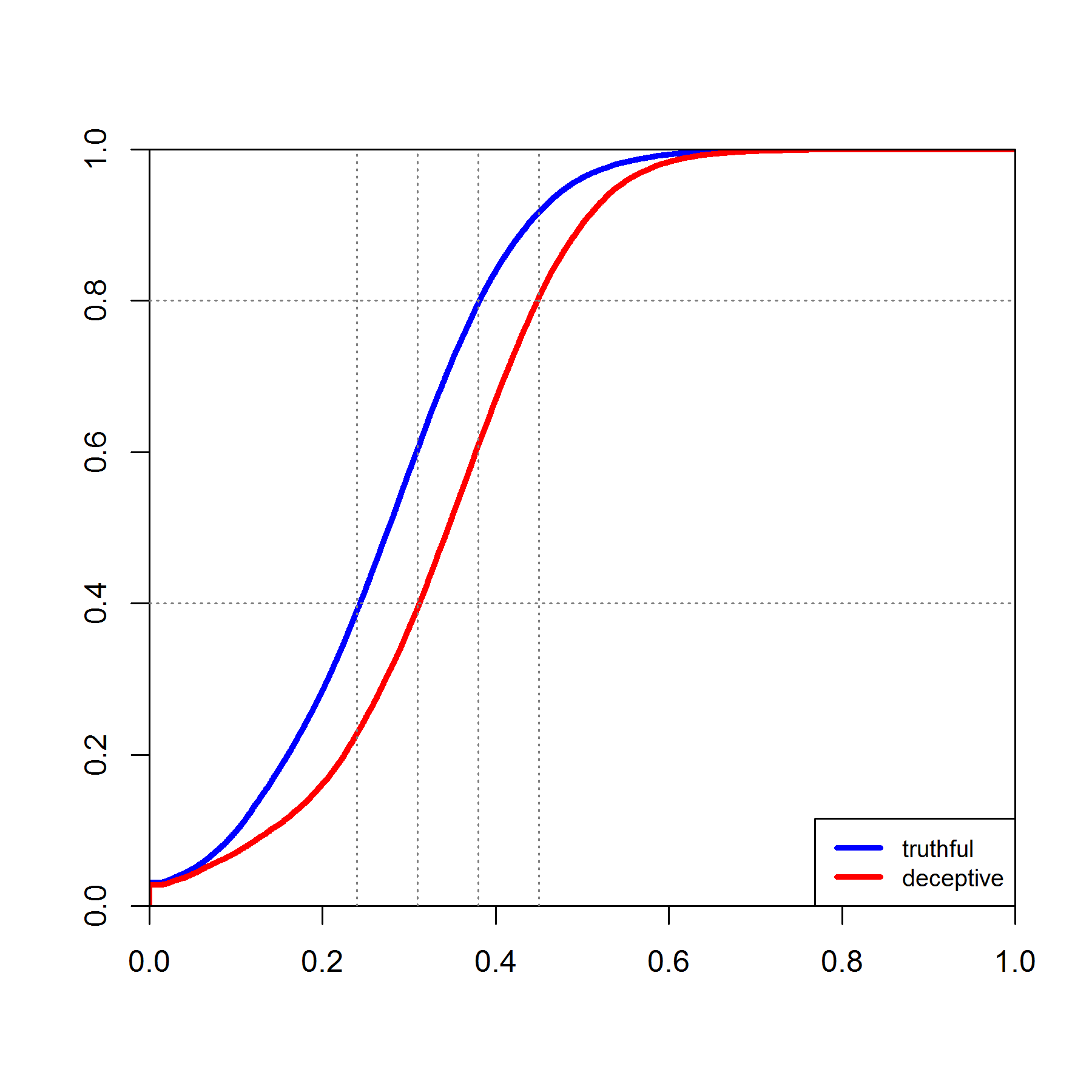}
\label{fig:cdf_ott_mihalcea}}
\subfloat[Subfigure 2 list of figures text][Cosine]{
\includegraphics[width=0.23\textwidth, trim = 9mm 15mm 8mm 15mm, clip]{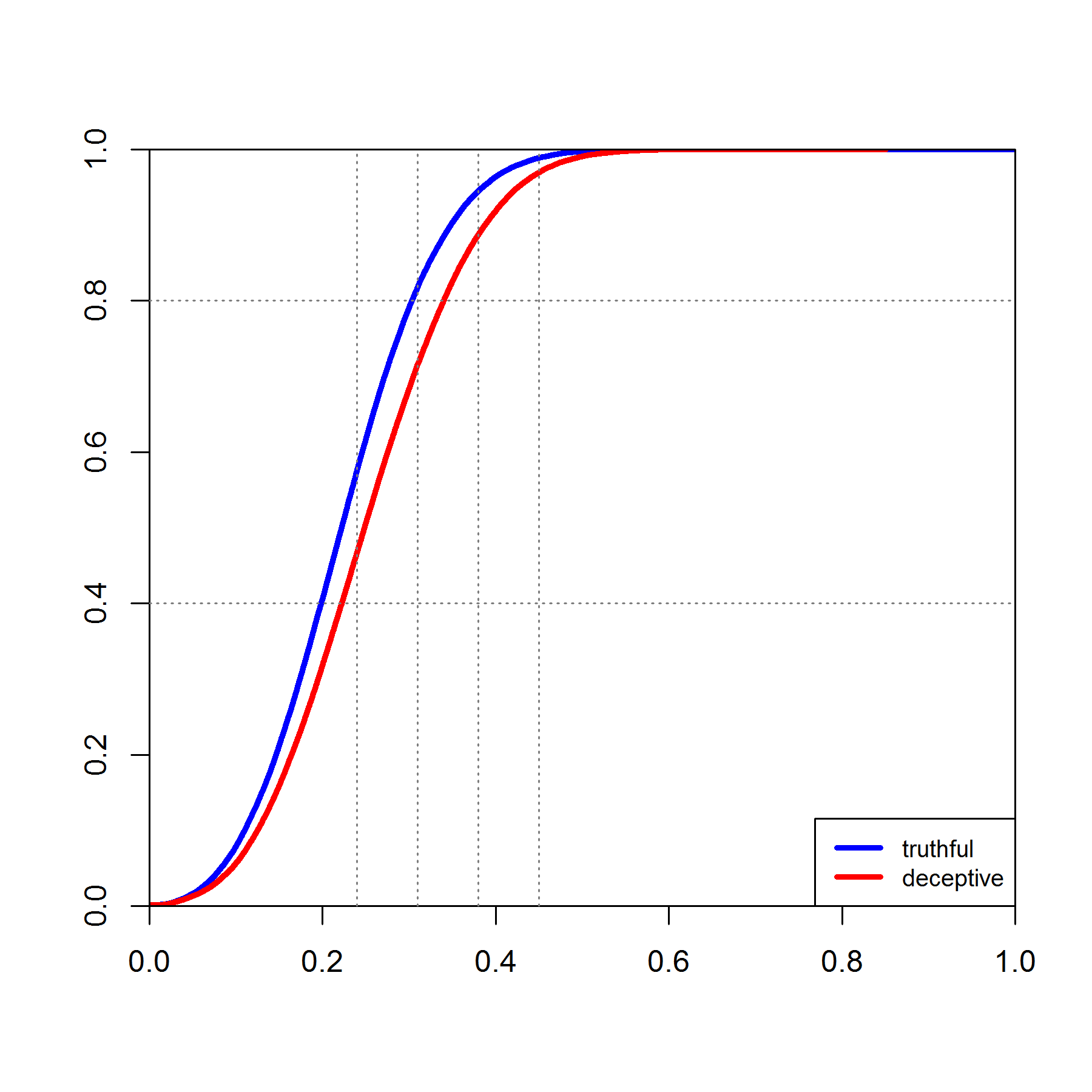}
\label{fig:cdf_ott_cosine}}
\caption{Ott dataset - cumulative percentage of fake reviews (red color) and truthful reviews (blue color) vs. similarity measures values}    
\label{fig:cdf_ott}
\end{figure}

We computed the pairwise similarity between truthful and fake reviews across the dataset and plotted the cumulative distribution functions (CDF) of each in Figure \ref{fig:cdf_ott}. They show the amount of content similarity for the truthful/fake reviews taken separately as well as the position and bounds for each type and the gaps between the two curves. Regardless of the type of similarity measure used, vectorial or semantic, the curves are clearly separated. For truthful reviews (blue), the curve appears towards the left of the plot, while for fake reviews (red) it is more to the right. This means that for a fixed cumulative percentage value, the similarity value \textit{$x_t$} for truthful reviews will be lower than for spam reviews \textit{$x_d$}. 

Figure \ref{fig:cdf_ott_cosine} shows that 80\% of truthful reviews are bounded by a cosine similarity value of 0.32, compared to the 0.34 for fake reviews. The difference is only of 2\% compared to the semantic similarity measure which showed a gap of 6\%.

\subsection{Bag-of-words LDA model}

\begin{figure}
\subfloat[Subfigure 1 list of figures text][Precision]{
\includegraphics[width=0.23\textwidth]{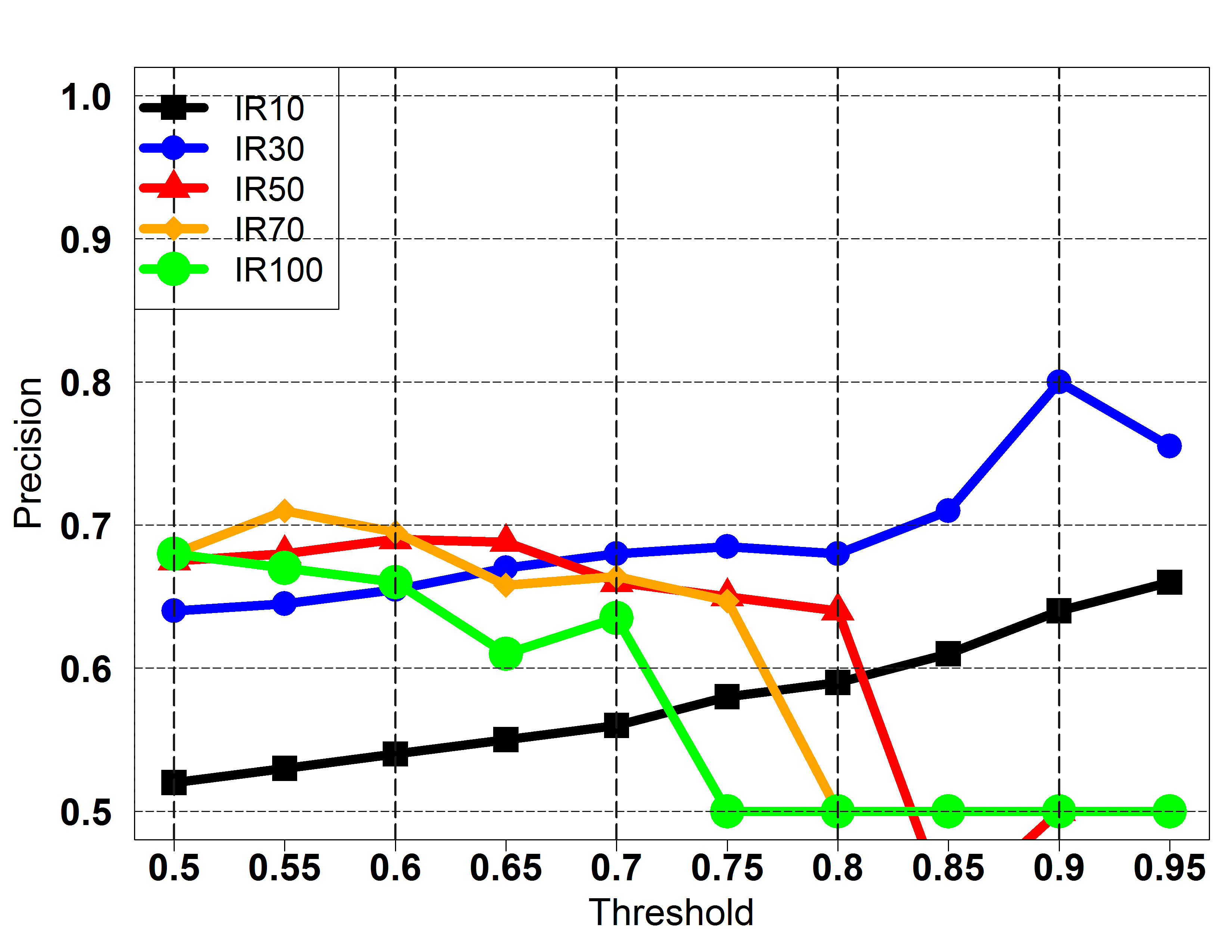}
\label{fig:yelp-lda-precision}}
\subfloat[Subfigure 2 list of figures text][F1 score]{
\includegraphics[width=0.23\textwidth]{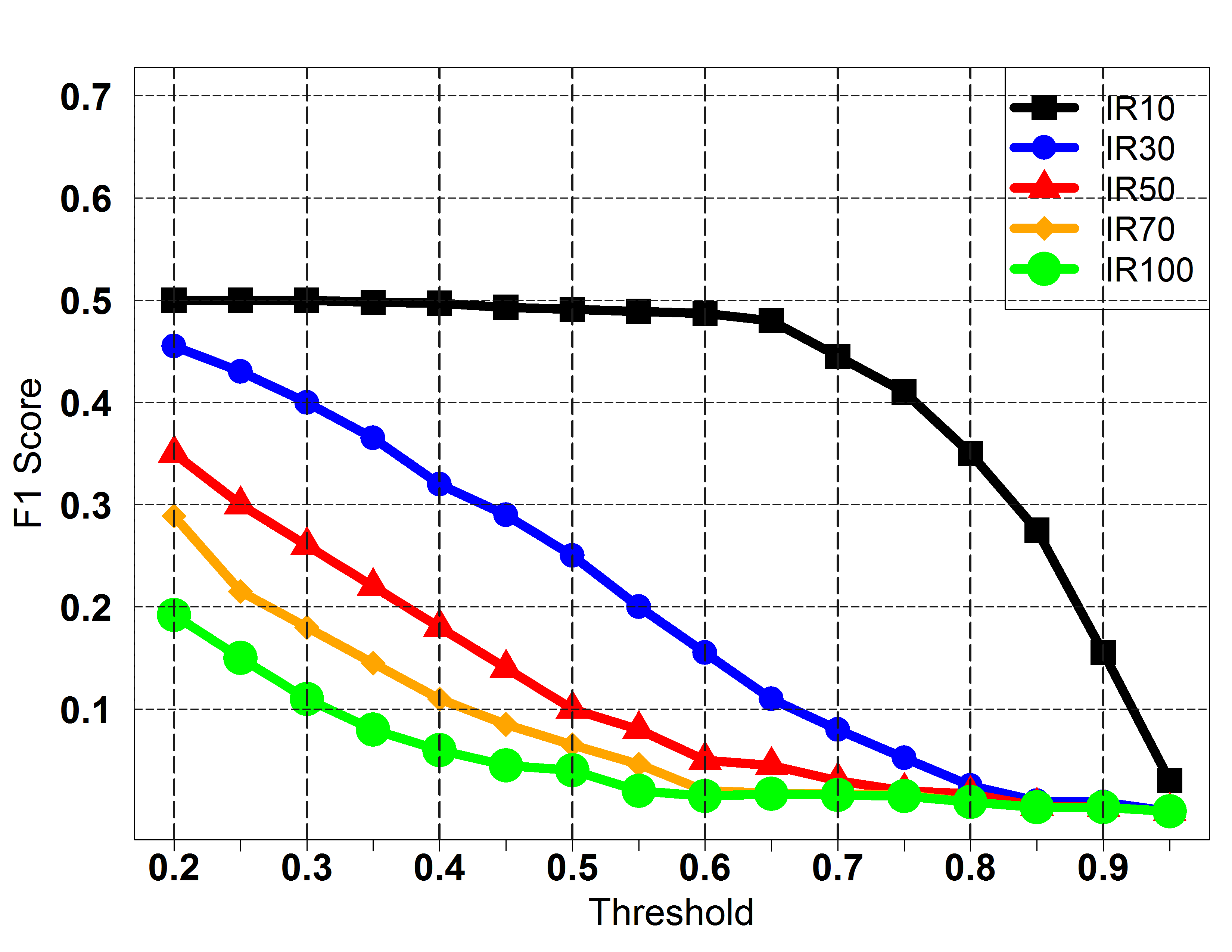}
\label{fig:yelp-lda-f1score}}
\caption{Yelp - classifier performance for IR similarity with bag-of-words}
\label{fig:yelp_lda}
\end{figure}

The classifier results for the Yelp dataset can be seen in Figure \ref{fig:yelp_lda}, where the results for the IR measure were plotted. IR10 refers to a LDA model with 10 topics, IR30 to a model with 30 topics and so on.

The classifier performs best in terms of precision for 30 topics for both Yelp and Trustpilot datasets. For Yelp IR30 has a smoother climb, reaching a precision of 65\% at a 0.6 threshold. It spikes at 80\% at a similarity threshold of 0.9. For 10 topics it climbs to a precision of above 60\%. The F1 score shows 10 topics are overall better than 30, but the difference in precision between the two curves is significant. It did not perform well when a larger number of topics was used.

\begin{figure}
\subfloat[Subfigure 1 list of figures text][Precision]{
\includegraphics[width=0.23\textwidth]{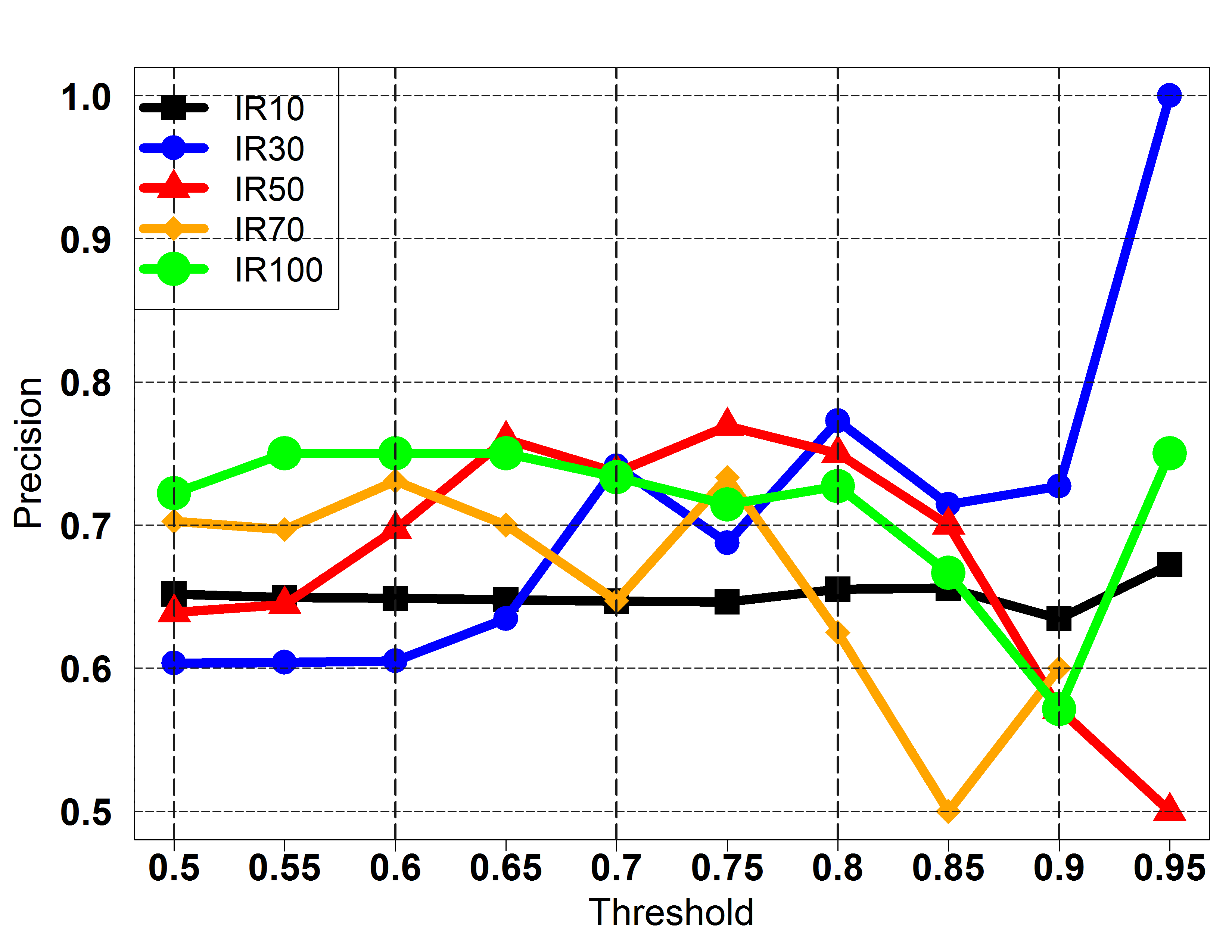}
\label{fig:trustpilot-lda-precision}}
\subfloat[Subfigure 2 list of figures text][F1 score]{
\includegraphics[width=0.23\textwidth]{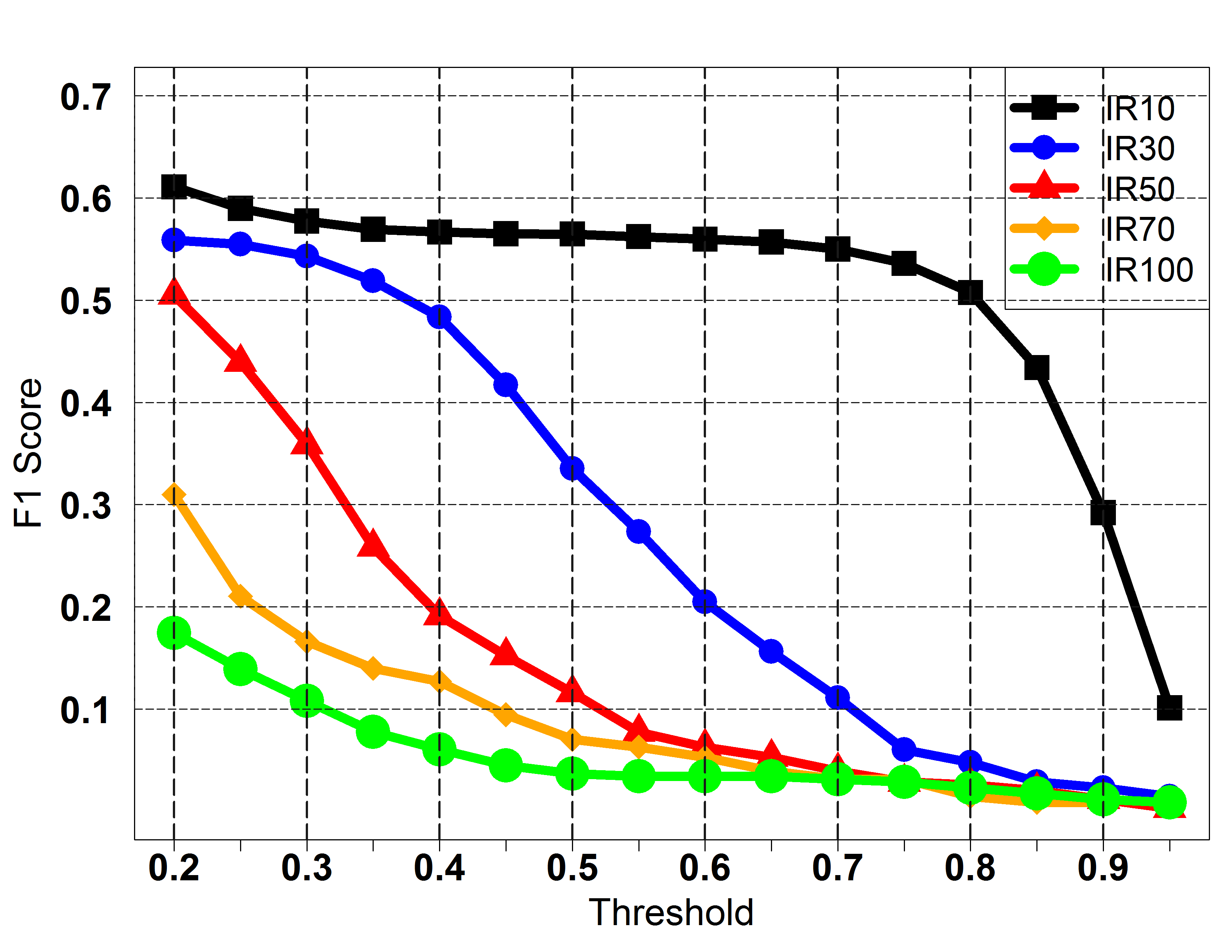}
\label{fig:trustpilot-lda-f1score}}
\caption{Trustpilot - classifier performance for IR similarity with bag-of-words}
\label{fig:trustpilot_lda}
\end{figure}

Figure \ref{fig:trustpilot-lda-precision} shows the precision according to each value of the spam threshold for the Trustpilot dataset. IR30 scored best and although the precision is not monotonic, generally it does not register significant drops as the threshold is increased. For thresholds of at least 0.7, it remains above 70\% and it peaks at 98\% for a 0.95 threshold. The recall and F1 score values are consistent in relation to the number of topics. As the number of topics is increased, the model performance decreases. For the Trustpilot dataset, the similarity results for 10 topics show the best F1 score, but the precision is more or less a flat line at 65\%. An explanation for this result could be that 10 topics are way too few to distinguish between honest reviewers and spammers.

\subsection{Bag-of-opinion-phrases LDA model}
For the bag-of-words approach, the classifier performed worse the more topics were used. However, for opinion phrases, using the Yelp dataset, there seems to be a smooth increase in performance as the similarity threshold increases, coupled with an increase in the number of topics. Intuitively, it makes more sense, since increasing the number of topics should create a better topics separation using opinion phrases. This causes reviews which mention the same aspects and sentiments to score higher in terms of their topic distributions similarity. The model performed badly on the Trustpilot dataset, giving more or less a flat precision regardless of the number of topics, therefore we did not plot the results. The poor performance could be a consequence of the dataset being much smaller than Yelp and plus, Trustpilot reviews are generally much shorter. Also opinion phrases induce topic sparseness even more than individual words. 

\begin{figure}
\subfloat[Subfigure 1 list of figures text][Precision]{
\includegraphics[width=0.23\textwidth]{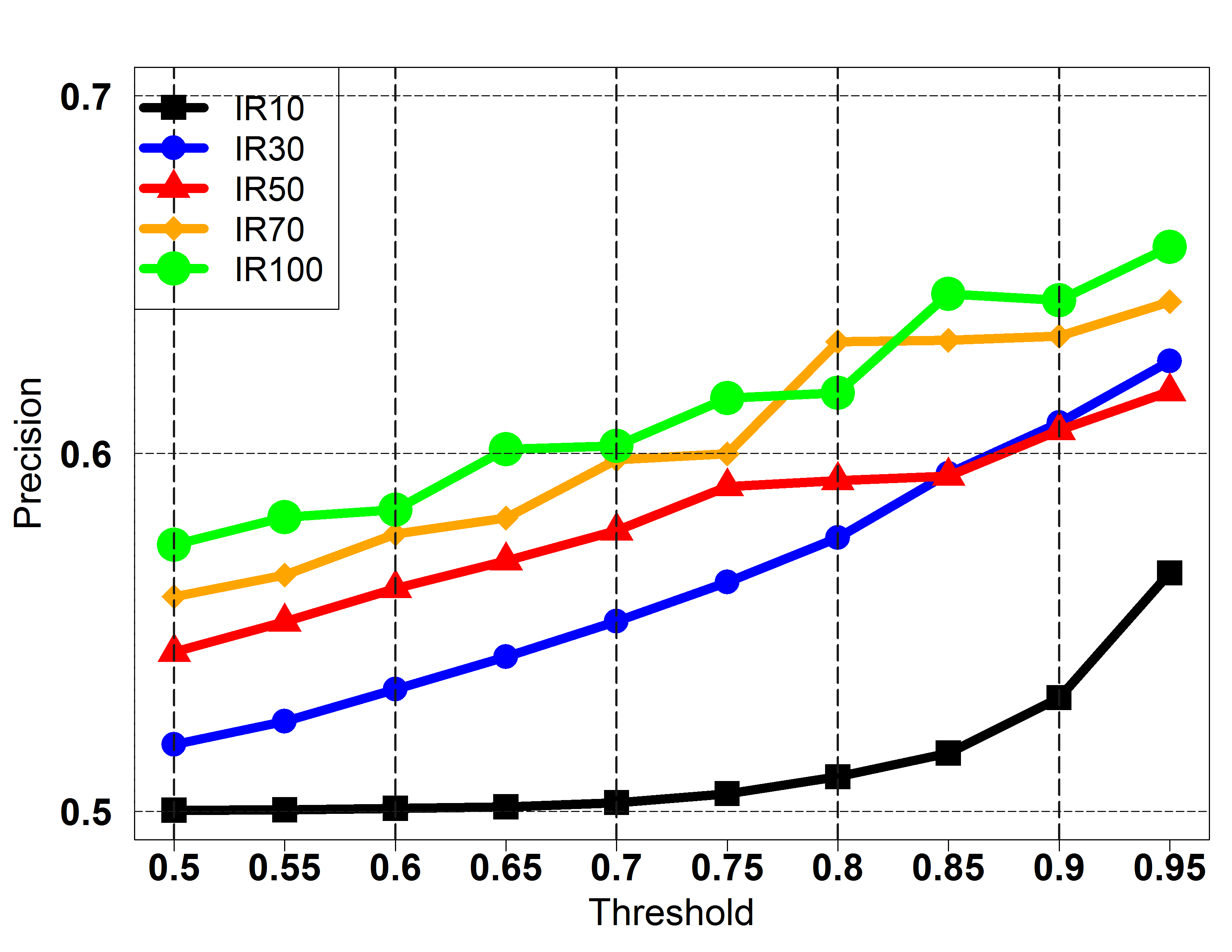}
\label{fig:yelp-lda-op-precision}}
\subfloat[Subfigure 2 list of figures text][F1 score]{
\includegraphics[width=0.23\textwidth]{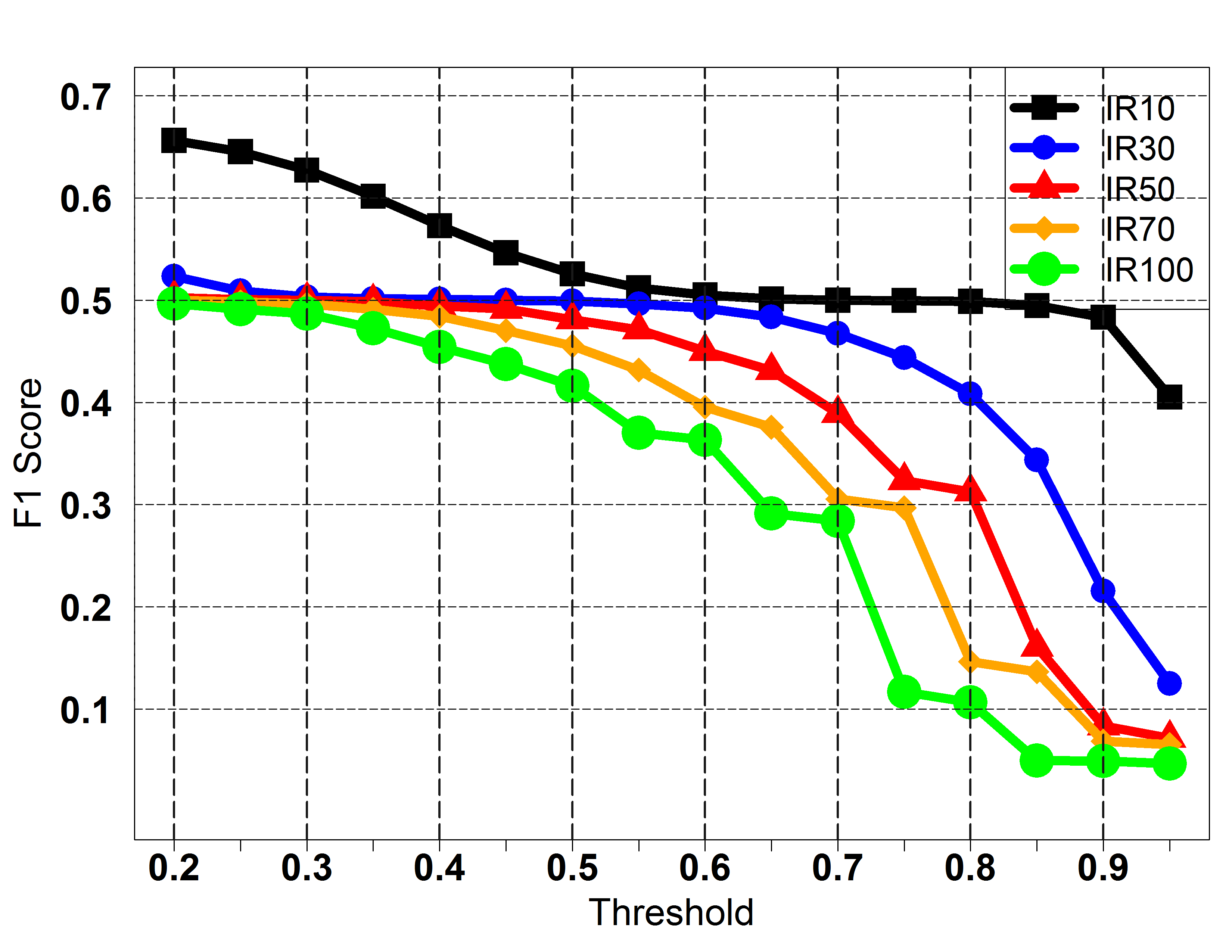}
\label{fig:yelp-lda-op-f1score}}
\caption{Yelp - classifier performance for IR similarity with bag-of-opinion-phrases}
\label{fig:yelp_lda_op}
\end{figure}

In the bag-of-words approach, the granularity of topics was not high enough, as both honest and spam authors tend to mention the same aspects about a business. The aspect-sentiment pair however improves on this granularity and increases the likelihood of reviews written by the same author to stick together. Figure \ref{fig:yelp-lda-op-precision} shows the classifier precision for 100 topics achieves 65\% for a 0.85 threshold, whereas for 30 or 50 topics, it is only close to 60\%. The recall is higher for 30 topics, as it would be expected.

The LDA models achieved a lower precision than the semantic and vectorial ones, but their main advantages are performance and being language agnostic, meaning they could infer review semantics regardless of the language. Although WordNets have been created for other languages besides English, none match the corpus scale of the English version.

\section{Conclusions}
We proposed two new approaches to the opinion spam detection problem. The first detection method is based on semantic similarity, which uses WordNet to compute the relatedness between words. Variants of the cosine similarity were also introduced, which together with the simple cosine measure were used as comparison baselines. Experimental results showed that semantic similarity can outperform the vectorial model in detecting spam reviews, capturing more subtle textual clues. The precision of the review classifier showed high results, and given the similarity threshold can be dynamically changed, it may make the method viable for a production detection system. 

We also proposed a method to detect opinion spam, using recent research aimed at extracting product aspects from short texts, such as user opinions and forums. We experimented with a bag-of-words model which performed well for only small number of topics. We experimented with a bag-of-opinion-phrases and the results showed a smooth increase in performance as the similarity threshold was increased, coupled with an increase in the number of topics. This also matched the intuition that the more topics we used, the less they would overlap and thus more reviews which mention the same aspects and sentiments would score higher in terms of their topic distributions similarity.

\bibliographystyle{abbrv}

\begin{thebibliography}{10}

\bibitem{blei:latent}
D.~M. Blei, A.~Y. Ng, and M.~I. Jordan.
\newblock Latent dirichlet allocation.
\newblock {\em Journal of Machine Learning Research}, 2003.

\bibitem{budanitsky:lexical}
A.~Budanitsky.
\newblock Lexical semantic relatedness and its application in natural language
  processing.
\newblock 1999.

\bibitem{dagan:similarity}
I.~Dagan, L.~Lee, and F.~C. Pereira.
\newblock Similarity-based models of word cooccurrence probabilities.
\newblock {\em Machine Learning}, 1999.

\bibitem{fei:exploiting}
G.~Fei, A.~Mukherjee, B.~Liu, M.~Hsu, M.~Castellanos, and R.~Ghosh.
\newblock Exploiting burstiness in reviews for review spammer detection.
\newblock In {\em AAAI}, 2013.

\bibitem{feng:distributional}
S.~Feng, L.~Xing, A.~Gogar, and Y.~Choi.
\newblock Distributional footprints of deceptive product reviews.
\newblock In {\em ICWSM}, 2012.

\bibitem{jindal:opinion}
N.~Jindal and B.~Liu.
\newblock Opinion spam and analysis.
\newblock In {\em WSDM}, 2008.

\bibitem{lim:detecting}
E.-P. Lim, V.-A. Nguyen, N.~Jindal, B.~Liu, and H.~W. Lauw.
\newblock Detecting product review spammers using rating behaviors.
\newblock In {\em CIKM}, 2010.

\bibitem{mihalcea:corpus}
R.~Mihalcea, C.~Corley, and C.~Strapparava.
\newblock Corpus-based and knowledge-based measures of text semantic
  similarity.
\newblock In {\em AAAI}, 2006.

\bibitem{moghaddam:onthedesign}
S.~Moghaddam and M.~Ester.
\newblock On the design of {LDA} models for aspect-based opinion mining.
\newblock In {\em CIKM}, 2012.

\bibitem{moghaddam:aspect}
S.~A. Moghaddam.
\newblock Aspect based opinion mining in online reviews.
\newblock Ph.D. thesis, Simon Fraser University, 2013.

\bibitem{mukherjee:spottingopinion}
A.~Mukherjee, A.~Kumar, B.~Liu, J.~Wang, M.~Hsu, M.~Castellanos, and R.~Ghosh.
\newblock Spotting opinion spammers using behavioral footprints.
\newblock In {\em KDD}, 2013.

\bibitem{mukherjee:spottingfake}
A.~Mukherjee, B.~Liu, and N.~Glance.
\newblock Spotting fake reviewer groups in consumer reviews.
\newblock In {\em WWW}, 2012.

\bibitem{mukherjee:fake}
A.~Mukherjee, V.~Venkataraman, B.~Liu, and N.~Glance.
\newblock Fake review detection: Classification and analysis of real and pseudo
  reviews.
\newblock 2013.

\bibitem{mukherjee:what}
A.~Mukherjee, V.~Venkataraman, B.~Liu, and N.~Glance.
\newblock What yelp fake review filter might be doing.
\newblock In {\em ICWSM}, 2013.

\bibitem{ott:finding}
M.~Ott, Y.~Choi, C.~Cardie, and J.~T. Hancock.
\newblock Finding deceptive opinion spam by any stretch of the imagination.
\newblock In {\em HLT}, 2011.

\bibitem{sandulescu:opinion}
V.~Sandulescu.
\newblock Opinion spam detection through semantic similarity.
\newblock MSc thesis (Unpublished), Technical University of Denmark, 2014.

\bibitem{toutanova:feature}
K.~Toutanova, D.~Klein, C.~D. Manning, and Y.~Singer.
\newblock Feature-rich part-of-speech tagging with a cyclic dependency network.
\newblock In {\em NAACL}, 2003.

\bibitem{xie:review}
S.~Xie, G.~Wang, S.~Lin, and P.~S. Yu.
\newblock Review spam detection via time series pattern discovery.
\newblock In {\em WWW Companion}, 2012.

\bibitem{zengin:user}
M.~Zengin and B.~Carterette.
\newblock User judgements of document similarity.
\newblock {\em Citeseer}, 2013.

\end{thebibliography}

\balancecolumns

\end{document}